\documentclass[]{bytedance_seed}

\usepackage[toc,page,header]{appendix}

\usepackage{minitoc}

\definecolor{cvprblue}{rgb}{0.21,0.49,0.74}
\usepackage{geometry}
\usepackage{booktabs}
\usepackage{multirow}
\usepackage{graphicx}
\usepackage[utf8]{inputenc}
\usepackage[most]{tcolorbox}
\usepackage{xcolor}

\newcommand{\ours}{\textsc{Inset}}

\definecolor{labelblack}{RGB}{0, 0, 0}
\definecolor{boxgray}{RGB}{242, 242, 242}
\definecolor{promptred}{RGB}{210, 0, 0}

\title{Images in Sentences: Scaling Interleaved Instructions for Unified Visual Generation}

\author{Yabo Zhang}
\author{Kunchang Li}
\author{Dewei Zhou}
\author{Xinyu Huang}
\author[\dagger]{Xun Wang}

\affiliation{ByteDance Seed}

\contribution[\dagger]{Project lead}

\abstract{
While recent advancements in multimodal language models have enabled image generation from expressive multi-image instructions, existing methods struggle to maintain performance under complex interleaved instructions. 
This limitation stems from the structural separation of images and text in current paradigms, which forces models to bridge difficult long-range dependencies to match descriptions with visual targets. 
To address these challenges, we propose \texttt{I}mages i\texttt{N} \texttt{SE}n\texttt{T}ences (\textit{a.k.a}, $\ours$), a unified generation model that seamlessly embeds images as native vocabulary within textual instructions. 
By positioning visual features directly at their corresponding semantic slots, $\ours$ leverages the contextual locality of transformers for precise object binding, effectively treating images as dense, expressive language tokens. 
Furthermore, we introduce a scalable data engine that synthesizes 15M high-quality interleaved samples from standard image and video datasets, utilizing VLMs and LLMs to construct rich, long-horizon sequences. 
Evaluation results on InterleaveBench demonstrate that $\ours$ significantly outperforms state-of-the-art methods in multi-image consistency and text alignment, with performance gaps widening as input complexity increases. 
Beyond standard generation, our approach inherently extends to multimodal image editing, integrating visual content as part of the instruction to facilitate highly expressive and creative visual manipulations.    
}

\date{\today}

\begin{document}
\maketitle

\begin{figure*}[t]
   \begin{center}
   \includegraphics[width=.95\linewidth]{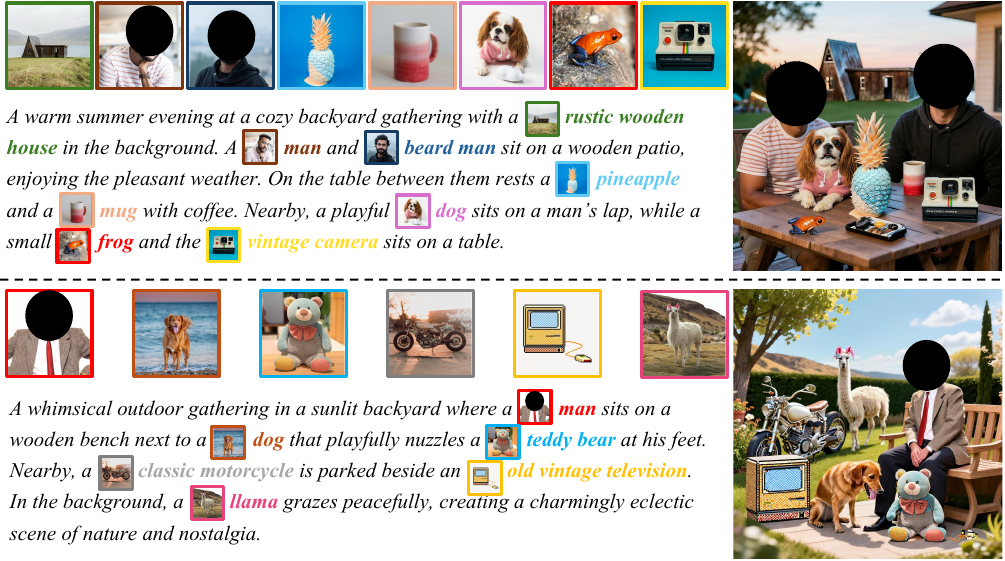}
   \end{center}
   \caption{
   \textbf{Showcases of INSET in Interleaved Image Generation and Editing.}
   By embedding images as native tokens within text, $\ours$ leverages contextual locality of transformers for precise object binding, enabling high-fidelity results in both complex generation and editing tasks.
   }
    \label{fig:method}
\end{figure*}

\section{Introduction}
Recent breakthroughs in multimodal understanding have revolutionized how models perceive and describe visual concepts~\cite{Qwen2.5-VL,gpt4,llava,comanici2025gemini,kosmos2,gpt4v}. 
This progress has propelled image generation beyond text-only prompts~\cite{rombach2022high,podell2023sdxl,esser2024scaling,flux2024}, embracing expressive interleaved image-text instructions~\cite{deng2025bagel,labs2025flux1kontextflowmatching,wu2025janus,ma2025janusflow,wu2025qwenimagetechnicalreport,wu2025omnigen2,zhou2024transfusion}. 
However, current methods fail to fully capitalize on this potential. 
Although capable of handling straightforward and few-reference scenarios, their performance drops sharply when they face complex multi-image constraints.

This inability to scale to complex scenarios stems from \textit{(i)} the indirect referencing mechanism and \textit{(ii)} the scarcity of complex interleaved data.
First, existing methods~\cite{deng2025bagel,labs2025flux1kontextflowmatching,wu2025qwenimagetechnicalreport,wu2025omnigen2,xie2025show,lin2025uniworld,seedream2025seedream,wang2024emu3} rely on an indirect query-based paradigm where visual content is retrieved via explicit indices, such as ``the dog in Image 1''. 
This design compels the model to simultaneously learn to align abstract indices with distant visual features and adjust their attributes and relationships based on the instruction.
Consequently, as input sequences lengthen with multiple reference images, the model often fails to accurately bind attributes to their corresponding targets, frequently neglecting specific image inputs.
Second, existing interleaved datasets~\cite{xia2025dreamomni2,xiao2025omnigen,ye2025echo,deng2025bagel} suffer from limited scale and complexity. 
Although they may include multiple reference images, the sequences are typically short and the interactions between text and images are rudimentary. 
They lack the rich, long-horizon interleaved examples necessary to teach the model how to handle intricate compositional reasoning involving dense visual contexts.

To overcome these challenges, we propose $\ours$, a unified generation model that seamlessly embeds images into sentences as native vocabulary, along with a scalable data engine.
Instead of treating images as external references requiring retrieval, we position visual features directly at their corresponding semantic slots within the instruction. 
Conceptually, $\ours$ regards input images as a detailed form of language, which broadens the input domain from text-only prompts to expressive interleaved instructions.
This interleaved architecture leverages the contextual locality of transformers~\cite{liu2024lost} to directly bind textual descriptions with visual targets, enabling the model to focus on comprehending the intricate interleaved inputs.
Furthermore, we develop a scalable data engine to construct high-quality interleaved data from standard image and video datasets. 
For static images, the data engine utilizes VLMs~\cite{guo2025seed1} to detect salient objects and generate granular descriptions, which are then synthesized by an LLM~\cite{guo2025seed1} into natural text sequences with visual embeddings explicitly placed at their semantic positions. 
Extending to video, it utilizes VLMs to establish object correspondence between frame pairs, prioritizing entities that undergo significant visual changes. 
These dynamic objects are then processed via the identical pipeline used for static images, explicitly enabling the model to learn how to manipulate visual states in response to textual instructions.

To comprehensively evaluate capabilities on complex interleaved tasks, we introduce InterleaveBench, a benchmark featuring multi-image compositions with intricate interleaved instructions. 
We implement $\ours$ on top of BAGEL and train it on 15M samples curated by our data engine. 
Experimental results demonstrate that $\ours$ surpasses all competing methods in multi-image consistency and significantly outperforms open-source models in text alignment. 
Notably, this performance advantage becomes increasingly pronounced as the number of input images grows, validating the scalability of our approach. 
Beyond generation, our interleaved format naturally extends to image editing, generalizing text-guided editing into a multimodal paradigm where both textual instructions and visual reference tokens guide the editing process.

Our contributions are summarized as follows:
\begin{itemize}
    \item We propose $\ours$, a unified generation model that embeds images as native vocabulary within instructions, utilizing the contextual locality to achieve precise object binding.
    
    \item We develop a scalable data engine that constructs 15M high-quality interleaved samples from image and video datasets, and introduce InterleaveBench for evaluating complex multi-image tasks.
    
    \item Experiments show that $\ours$ achieves superior performance in image and text consistency, with advantages amplifying as complexity increases, and naturally generalizes to multimodal image editing.
\end{itemize}
\section{Related Works}
\subsection{Unified Image Generation Models}
Following the success of text-to-image models~\cite{rombach2022high,podell2023sdxl,flux2024}, research has increasingly focused on enabling interleaved image-text inputs~\cite{chen2025janus,deng2025bagel,karypis1999chameleon,labs2025flux1kontextflowmatching,ma2025janusflow,zhou2024transfusion,lin2025uniworld,sun2024generative,xia2025dreamomni2,seedream2025seedream,wei2025skywork,wang2025skywork}. 
Early attempts~\cite{wei2023elite,wang2024instantid,ye2023ip} primarily relied on pre-trained image encoders such as CLIP~\cite{radford2021learning} to extract visual features, but are prone to rigid copy-paste artifacts and often conflate features when processing multiple reference images. 
With the rapid advancement of multimodal large language models, recent paradigms have shifted towards leveraging these powerful understanding models to handle multimodal inputs. 
Among these, autoregressive models~\cite{wang2025simplear,sun2024generative,karypis1999chameleon} adopt discrete image tokenization for unified modeling, though their quality is often bottlenecked by the visual tokenizer.
~\cite{xie2024show,zhou2024transfusion,xiao2025omnigen} employ a single transformer for both modalities, yet often trail behind specialized models in generation fidelity. 
Consequently, the majority of recent works~\cite{pan2023kosmos,wu2025qwenimagetechnicalreport,wu2025janus,deng2025bagel,labs2025flux1kontextflowmatching,lin2025uniworld,seedream2025seedream,wu2025harmonizing,seedream2025seedream,liao2025mogao,chen2025blip3} adopt a hybrid strategy that connects understanding and generation modules without sharing parameters, allowing the generator to benefit from MLLM capabilities. 
Despite these progressions, current multimodal generation models have not fully unlocked the potential of advanced understanding models, showing competence in simple interleaved inputs but often faltering when facing complex, multi-step instructions.

\subsection{Interleaved Image-Text Datasets} 
The availability of high-quality interleaved datasets is pivotal for advancing multimodal generation, yet existing options face significant limitations in supporting complex instruction-following. 
Large-scale web-crawled corpora~\cite{li2024omnicorpus,zhu2023multimodal} often suffer from loose semantic alignment and noisy text-image correlations, rendering them suboptimal for precise generation tasks. 
Conversely, datasets derived from video sequences~\cite{deng2025bagel} are primarily tailored for multi-turn editing with high visual redundancy, lacking the capacity to chain distinct visual concepts.
Subject-driven collections (\textit{e.g.}, X2I-subject~\cite{xiao2025omnigen}) are typically constrained by limited input images and simplistic commands. 
More recently, synthetic datasets~\cite{ye2025echo,xia2025dreamomni2,mou2025dreamo} have utilized generative models for data construction. 
However, these approaches struggle to maintain diversity at scale and are inherently bottlenecked by the capabilities of the source generative models. 
To bridge this gap, we introduce a scalable data engine designed to construct rich, complex interleaved sequences derived from real-world scenarios, ensuring both diversity and semantic precision.
\section{Method}
\label{sec:method}
\begin{figure*}[t]
   \begin{center}
   \includegraphics[width=.99\linewidth]{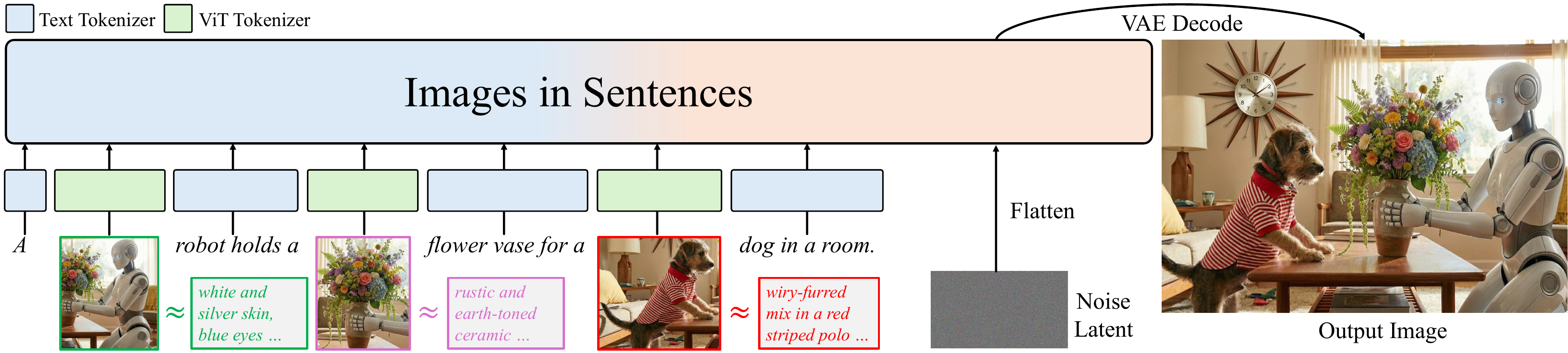}
   \end{center}
   \caption{
   \textbf{Overview of INSET.}
   Our method positions visual features directly at their corresponding semantic slots within the text instruction. 
   By using a semantic ViT tokenizer as vision encoder, it treats input images as a detailed form of language, expanding text-only prompts into expressive interleaved instructions.
   }
    \label{fig:method}
\end{figure*}
\begin{figure*}[t]
   \begin{center}
   \includegraphics[width=.99\linewidth]{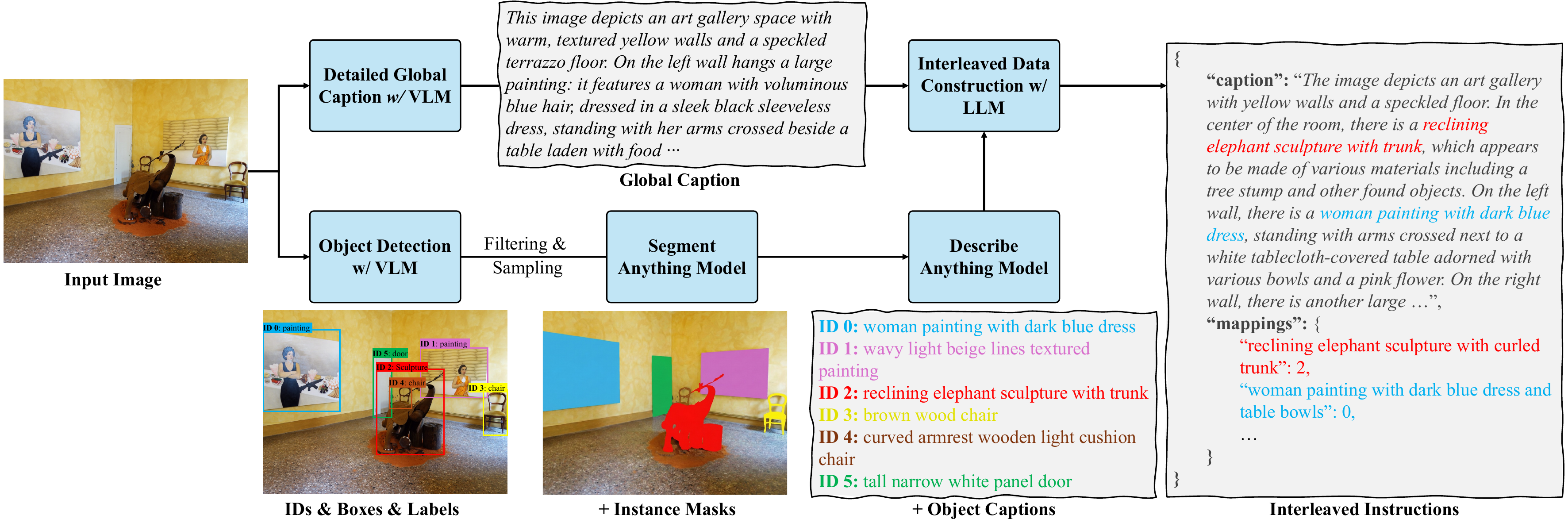}
   \end{center}
   \caption{
   \textbf{Synthesizing Interleaved Data from Images.}
   We synthesize training data by \textit{(i)} generating a global narrative via VLM, \textit{(ii)} extracting fine-grained instance masks and object captions, and \textit{(iii)} employing an LLM to weave these visual instances into their precise semantic contexts. This process transforms static images into expressive interleaved instructions with structured text-image mappings.
   }
    \label{fig:image_pipeline}
\end{figure*}

In this section, we present $\ours$, a unified framework designed to master complex multi-image generation through a native interleaved formulation. 
We begin by detailing the modeling paradigm in Sec.~\ref{sec:modeling}, which embeds images directly as vocabulary within instructions to ensure precise semantic binding. 
To support this approach, we introduce a scalable data engine in Sec.~\ref{sec:data_engine} that curates 15M high-quality interleaved samples from real-world image and video corpora. 
Finally, in Sec.~\ref{sec:bench}, we propose InterleaveBench, a rigorous benchmark and evaluation protocol tailored for assessing complex interleaved scenarios.

\subsection{Unified Interleaved Modeling}
\label{sec:modeling}

\noindent \textbf{Native Interleaved Formulation.}
Existing unified generation models predominantly rely on an indirect query-based paradigm, where visual content is retrieved via explicit indices.
For instance, given the inputs in Figure~\ref{fig:method}, these models typically segregate reference images from the textual instruction (e.g., \texttt{[Image1][Image2][Image3]} + \texttt{"A robot in image 1 holds a flower vase from image 2..."}).
This design compels the model to contend with long-range dependencies between the textual instruction and distant visual features.
Consequently, as input sequences lengthen with multiple reference images, the model often fails to accurately bind attributes to their corresponding targets or simply neglects specific image inputs.

In contrast, as illustrated in Figure~\ref{fig:method}, $\ours$ conceptually regards input images as a detailed form of language, seamlessly embedding them into sentences as native vocabulary.
This formulation broadens the input domain from simple text-only prompts to expressive interleaved instructions.
By positioning visual features directly at their corresponding semantic slots (e.g., \texttt{"A [Image1] robot holds a [Image2] flower vase..."}), we leverage the inherent contextual locality of transformers to directly bind textual descriptions with visual targets.
This explicit alignment relieves the model from the struggle of long-range dependency resolution, enabling it to focus entirely on comprehending and executing intricate interleaved instructions.

\noindent \textbf{Model Architecture.}
Following BAGEL, $\ours$ adopts a Mixture-of-Transformer architecture, including an understanding branch designed to process interleaved image-text instructions and a generation branch dedicated to image synthesis.
Diverging from standard dual-feature inputs, we only input semantic ViT embeddings and discard pixel-level VAE latent features.
In multi-image scenarios, the inclusion of VAE latents often biases the model towards ``image-pasting'' issues, where reference objects are rigidly copied rather than semantically integrated.
By relying solely on ViT features, we mitigate this trivial copying and encourage the model to perform deeper semantic reasoning for consistent composition.

\noindent \textbf{Inference Strategy.}
During inference, the visual modality tends to dominate the generation process, often overshadowing textual instructions.
To rectify this imbalance, we adopt a two-stage guidance strategy.
First, we calibrate the interplay between modalities by boosting the text influence relative to a visual-only baseline.
Second, we apply classifier-free guidance using the null embedding $\emptyset$ as the uncondition input.
Formally, let $\mathbf{c}_t$ and $\mathbf{c}_v$ denote the text and visual conditions, and $\emptyset$ represent the null token for missing modalities.
We use $s_1$ and $s_2$, to control the text-image balance and the overall generation strength, respectively.
The balanced conditional estimate $\hat{\epsilon}_{\text{bal}}$ and the final noise prediction $\tilde{\epsilon}_\theta$ are computed as:
\begin{equation}
    \hat{\epsilon}_{\text{bal}} = \epsilon_\theta(z_t, \emptyset, \mathbf{c}_v) + s_1 \cdot (\epsilon_\theta(z_t, \mathbf{c}_t, \mathbf{c}_v) - \epsilon_\theta(z_t, \emptyset, \mathbf{c}_v)),
\end{equation}
\begin{equation}
    \tilde{\epsilon}_\theta = \epsilon_\theta(z_t, \emptyset, \emptyset) + s_2 \cdot (\hat{\epsilon}_{\text{bal}} - \epsilon_\theta(z_t, \emptyset, \emptyset)).
\end{equation}
By setting $s_1 = 4.0$, we explicitly enhance the adherence to textual descriptions before applying the global guidance scale $s_2=1.5$.

\subsection{Scalable Interleaved Data Engine}
\label{sec:data_engine}
To fully realize the potential of $\ours$, diverse and high-quality data is indispensable.
Addressing the critical scarcity of such resources, we propose a scalable interleaved data engine that autonomously mines and structures complex interleaved sequences directly from large-scale real-world image and video corpora.

\vspace{2pt}
\noindent \textbf{Synthesizing Interleaved Data from Images.}
To construct training data that mirrors the complexity of natural interleaved instructions, our pipeline seamlessly embeds visual instances into their precise semantic contexts, as illustrated in Figure~\ref{fig:image_pipeline}. The process comprises three stages:

\text{(1) Global Captioning.}
We first employ a VLM (\textit{e.g.}, Doubao-Seed-1.6-Vision) to generate a comprehensive global description of the image. This provides a narrative backbone, capturing the scene's overall context and spatial relationships.

\text{(2) Fine-grained Object Processing.}
Parallel to global captioning, we extract dense visual details. We utilize a VLM for object detection to obtain bounding boxes and category labels. Following a filtering and sampling step to remove low-quality candidates (\textit{e.g.}, extreme sizes), we apply the Segment Anything Model (SAM)~\cite{kirillov2023segment} to generate pixel-perfect instance masks. Finally, the Describe Anything Model (DAM)~\cite{lian2025describe} produces detailed object captions for each valid instance.

\text{(3) LLM-driven Interleaved Construction.}
In the final stage, an LLM synthesizes the interleaved instruction. Taking the global caption and the set of object triplets (label, mask, object caption) as input, the LLM rewrites the narrative to naturally incorporate the detected objects. It compresses detailed regional descriptions into concise descriptive phrases and outputs a structured JSON containing the final interleaved caption and a precise mapping between these textual phrases and their corresponding visual indices.
Through this pipeline, we curate 10M complex samples, each containing 3--8 input images, providing a dense signal for learning text-image correspondence.

\vspace{2pt}
\noindent \textbf{Synthesizing from Videos.}
Relying solely on static imagery risks training the model to merely ``copy-paste'' reference objects without adaptation.
To empower the model with dynamic state manipulation capabilities, we extend our data engine to video corpora.
Our goal is to leverage temporal changes to construct training pairs where the visual reference (from a source frame) and the generation target (a target frame) depict the same entity in distinctly different states.

\text{(1) Long-range Object Correspondence.}
We select frame pairs separated by distinct temporal intervals to maximize visual variance. Instead of relying on traditional tracking which struggles with large gaps, we concatenate both frames and feed them into a VLM. The VLM is prompted to jointly identify and match identical entities across the two views, ensuring robust correspondence even under significant view changes.

\text{(2) Dynamic State Filtering.}
To ensure the model learns transformation rather than reconstruction, we apply a dual-stage filter to select objects that undergo meaningful changes.
We first discard static pairs using ORB feature matching (high similarity).
Subsequently, we employ a \textit{lightweight} VLM (\textit{e.g.}, Doubao-Seed-1.6-Flash) to verify that the remaining pairs exhibit significant semantic alterations in action, pose, or morphology.

\text{(3) Cross-Frame Instruction Synthesis.}
We construct interleaved instructions specifically for the target frame.
Crucially, the visual tokens embedding in the instruction are cropped from the source frame.
During training, the model learn to preserve the object's identity provided by the source visual token, while simultaneously transforming its state (\textit{e.g.}, pose, lighting) to align with the textual description of the target frame.
This strategy yields 5M video-derived samples, explicitly training the model to manipulate object states according to textual instructions.

\subsection{InterleaveBench Construction}
\label{sec:bench}

Existing benchmarks, such as DreamBench++~\cite{peng2024dreambench++} and OmniContext~\cite{wu2025omnigen2}, often lack the complexity required for robust evaluation due to their limited reference images and simple spatial relationships.
To address this gap, we introduce \textbf{InterleaveBench}, a rigorous benchmark designed for complex multi-image scenarios.

\noindent \textbf{Dataset Curation.}
We source high-quality reference entities from DreamBench++~\cite{peng2024dreambench++}. For each test case, we sample $N \in [2, 5]$ distinct images and employ a VLM to filter for semantic compatibility.
We then generate intricate interleaved instructions that mandate logical spatial reasoning and adaptive attribute modification, rather than simple composition.
To ensure quality, all samples undergo rigorous human verification to filter out unnatural or conflicting prompts.

\noindent \textbf{Evaluation Protocol.}
Conventional metrics relying on holistic embeddings, such as CLIP~\cite{radford2021learning} or DINO~\cite{caron2021emerging}, struggle to accurately assess identity preservation within complex interleaved instructions involving multiple subjects.
To strictly quantify performance beyond these limitations, we implement a dual-perspective LLM-as-Judge framework.
\textit{(i) Image Consistency} evaluates identity preservation by assigning a rating on a 1--5 scale, which is subsequently normalized to the interval $[0, 1]$.
Crucially, it is designed to penalize fundamental identity drift while explicitly tolerating reasonable instruction-driven variations (\textit{e.g.}, pose or lighting changes), a nuance often misjudged by simple embedding distances.
\textit{(ii) Text Consistency} measures semantic alignment via a VQA-based approach~\cite{peng2024dreambench++}.
We leverage an LLM to pre-formulate a set of binary questions targeting specific attributes and relationships defined in the instruction.
During evaluation, these pre-defined questions are answered by a VLM to calculate an adherence score.
\section{Experiments}
\begin{table*}[t]
\centering
\caption{\textbf{Quantitative Comparisons Different Numbers of Input Images.}
Despite its smaller parameter scale, $\ours$ significantly outperforms all open-source methods and achieves performance comparable to closed-source models. As the number of input images increases, $\ours$ shows widening leads in both image and text consistency.
}

\label{tab:main_com}
\setlength{\tabcolsep}{1.5pt}
\resizebox{\textwidth}{!}{%
\begin{tabular}{lcccccccccccc}
\toprule
\multirow{2}{*}{\textbf{Method}} & \multirow{2}{*}{\textbf{\#Params.}} & \multicolumn{5}{c}{\textbf{Image Consistency}} & \multicolumn{5}{c}{\textbf{Text Consistency}} \\
\cmidrule(lr){3-7} \cmidrule(lr){8-12}
& & {Two Obj.} & {Three Obj.} & {Four Obj.} & {Five Obj.} & {Overall} & {Two Obj.} & {Three Obj.} & {Four Obj.} & {Five Obj.} & {Overall} \\
\midrule
\multicolumn{12}{c}{{\textit{Close-sourced Models}}} \\
\midrule
GPT-4o &- & 0.92 & 0.93 & 0.90 & 0.89 & 0.91 & 0.86 & 0.85 & 0.78 & 0.81 & 0.81 \\
Nano Banana &- & 0.92 & 0.94 & 0.83 & 0.83 & 0.88 & 0.83 & 0.87 & 0.74 & 0.76 & 0.79 \\
Seedream 4.0 &- & 0.90 & 0.93 & 0.91 & 0.91 & 0.91 & 0.84 & 0.88 & 0.85 & 0.85 & 0.85 \\
\midrule
\multicolumn{12}{c}{{\textit{Open-sourced Models}}} \\
\midrule
DreamOmni2 &7B+5B+12B & 0.83 & 0.80 & 0.69 & 0.65 & 0.75 & 0.71 & 0.67 & 0.51 & 0.47 & 0.57 \\
FluxKontext &5B+12B & 0.83 & 0.80 & 0.69 & 0.60 & 0.74 & 0.72 & 0.68 & 0.50 & 0.42 & 0.56 \\
Qwen Image &7B+20B & 0.85 & 0.75 & 0.70 & 0.41 & 0.69 & 0.78 & 0.66 & 0.48 & 0.20 & 0.49 \\
BAGEL &7B+7B & 0.68 & 0.62 & 0.57 & 0.57 & 0.61 & 0.59 & 0.48 & 0.40 & 0.38 & 0.45 \\
\midrule
\multirow{2}{*}{$\ours$ (Ours)} & \multirow{2}{*}{7B+7B} & \textbf{0.93} & \textbf{0.94} & \textbf{0.90} & \textbf{0.94} & \textbf{0.93} & \textbf{0.82} & \textbf{0.78} & \textbf{0.72} & \textbf{0.71} & \textbf{0.75} \\[-0.2em]
 & & \textcolor[rgb]{0.0, 0.5, 0.0}{(+0.08)} & \textcolor[rgb]{0.0, 0.5, 0.0}{(+0.14)} & \textcolor[rgb]{0.0, 0.5, 0.0}{(+0.20)} & \textcolor[rgb]{0.0, 0.5, 0.0}{(+0.29)} & \textcolor[rgb]{0.0, 0.5, 0.0}{(+0.18)} & \textcolor[rgb]{0.0, 0.5, 0.0}{(+0.04)} & \textcolor[rgb]{0.0, 0.5, 0.0}{(+0.10)} & \textcolor[rgb]{0.0, 0.5, 0.0}{(+0.21)} & \textcolor[rgb]{0.0, 0.5, 0.0}{(+0.24)} & \textcolor[rgb]{0.0, 0.5, 0.0}{(+0.18)} \\
\bottomrule
\end{tabular}%
}
\end{table*}
\begin{figure*}[ht]
   \begin{center}
   \includegraphics[width=.95\linewidth]{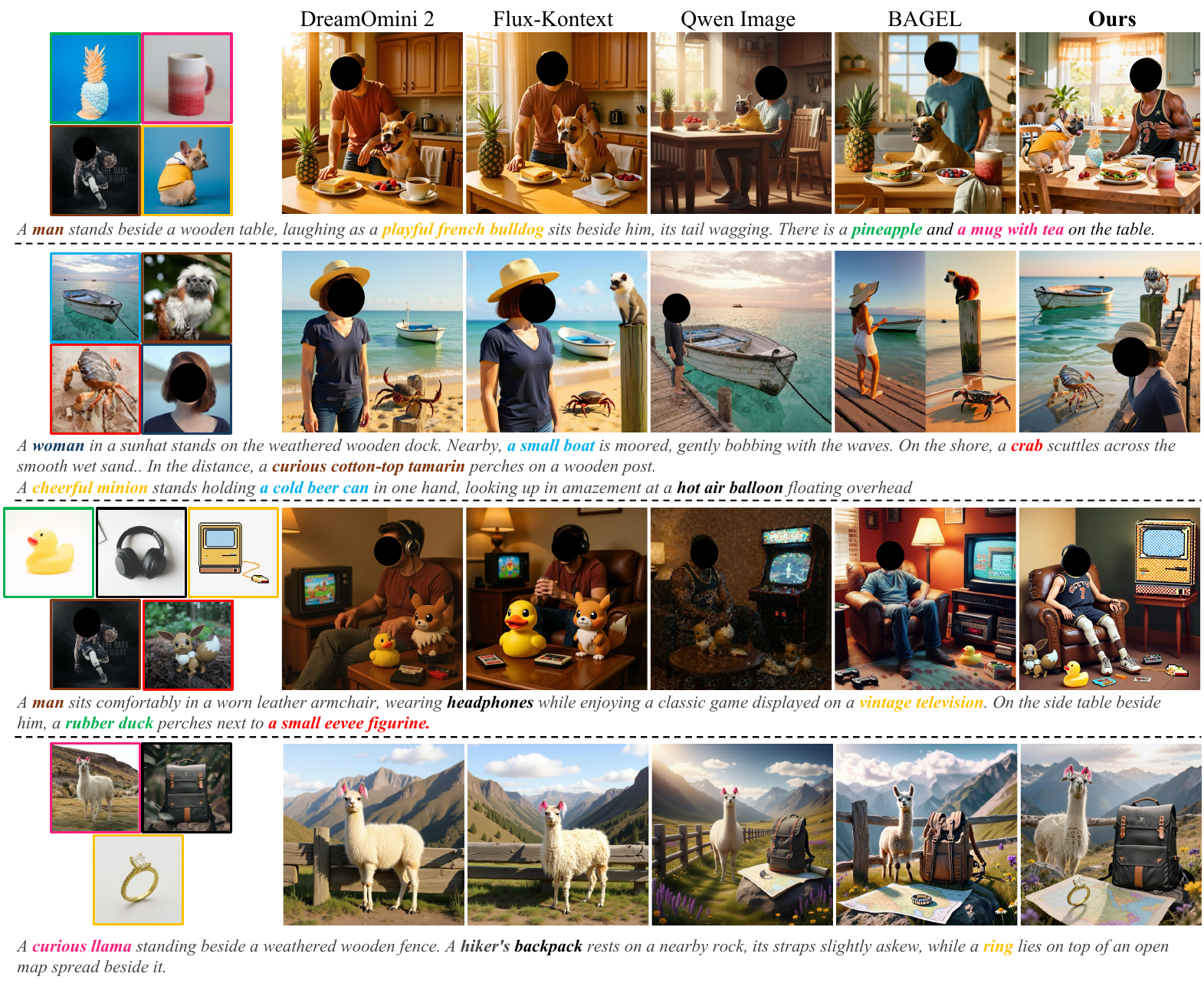}
   \end{center}
   \caption{
   \textbf{Qualitative Comparisons with the SOTA Open-sourced Methods.}
   $\ours$ significantly outperforms open-source baselines in visual consistency and precise attribute binding, avoiding common failure cases such as object misalignment or the inability to render specific actions and textures.
   }
    \label{fig:main_com_open}
\end{figure*}
\begin{figure*}[ht]
   \begin{center}
   \includegraphics[width=.95\linewidth]{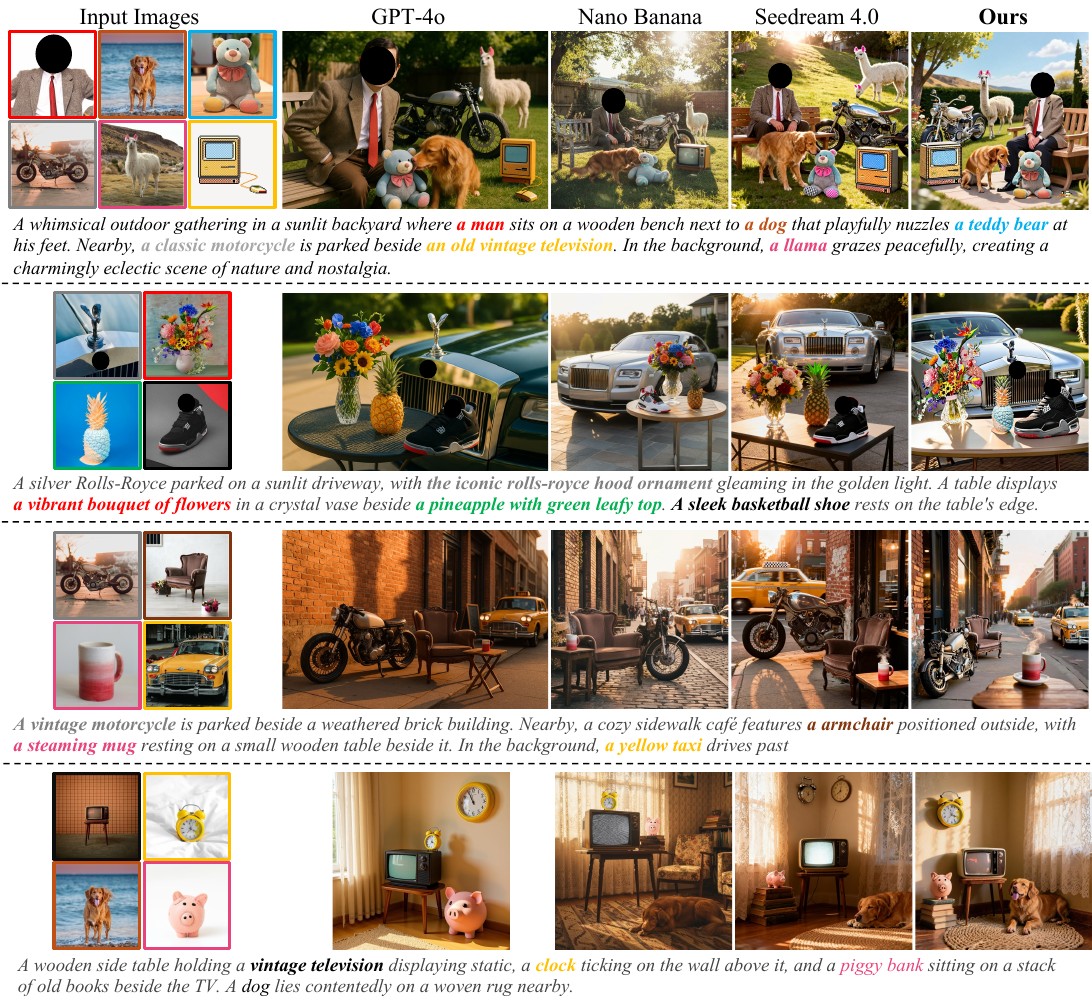}
   \end{center}
   \caption{
   \textbf{Qualitative Comparisons with Advanced Close-sourced Methods.}
   $\ours$ demonstrates superior image fidelity and stability when handling complex interleaved instructions, particularly in maintaining identity consistency across challenging visual contexts.
   }
    \label{fig:main_com_close}
\end{figure*}
\begin{figure}[ht]
   \begin{center}
   \includegraphics[width=.8\linewidth]{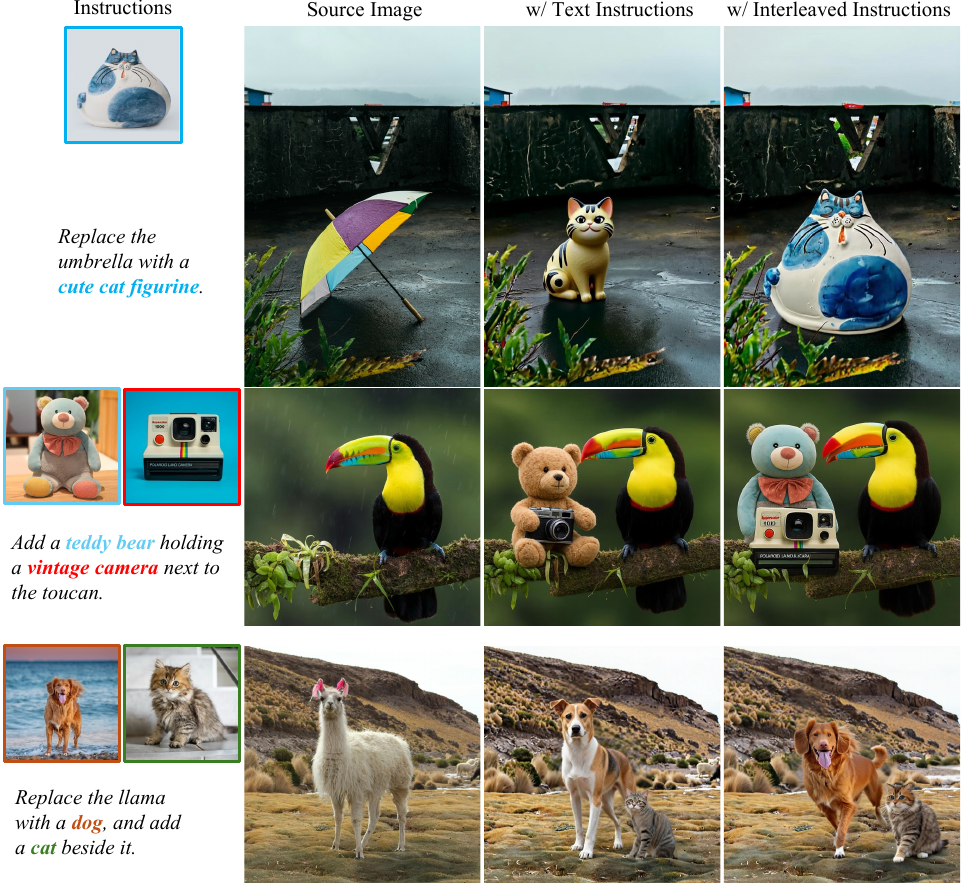}
   \end{center}
   \caption{
   \textbf{Emerging Multimodal Image Editing via Interleaved Instructions.}
   }
    \label{fig:main_edit}
\end{figure}
\begin{table*}[t]
\centering
\caption{
\textbf{Quantitative Ablation Study of Key Model Components.}
}
\label{tab:main_ablation}
\setlength{\tabcolsep}{1.5pt}
\resizebox{\textwidth}{!}{%
\begin{tabular}{lcccccccccc}
\toprule
\multirow{2}{*}{\textbf{Method}} & \multicolumn{5}{c}{\textbf{Image Consistency}} & \multicolumn{5}{c}{\textbf{Text Consistency}} \\
\cmidrule(lr){2-6} \cmidrule(lr){7-11}
 & {Two Obj.} & {Three Obj.} & {Four Obj.} & {Five Obj.} & {Overall} & {Two Obj.} & {Three Obj.} & {Four Obj.} & {Five Obj.} & {Overall} \\
\midrule
Baseline & 0.68 & 0.62 & 0.57 & 0.57 & 0.61 & 0.59 & 0.48 & 0.40 & 0.38 & 0.45 \\
Image First & 0.88 & 0.89 & 0.83 & 0.82 & 0.86 & 0.73 & 0.70 & 0.64 & 0.60 & 0.66 \\
\textit{w/o} Video-based Data & 0.90 & 0.92 & 0.91 & 0.91 & 0.91 & 0.68 & 0.65 & 0.56 & 0.57 & 0.60 \\
\textit{w/} VAE Feature & 0.91 & 0.90 & 0.74 & 0.68 & 0.82 & 0.74 & 0.71 & 0.46 & 0.42 & 0.56 \\
$\ours$ (Ours) & \textbf{0.93} & \textbf{0.94} & \textbf{0.90} & \textbf{0.94} & \textbf{0.93} & \textbf{0.82} & \textbf{0.78} & \textbf{0.72} & \textbf{0.71} & \textbf{0.75} \\
\bottomrule
\end{tabular}%
}
\end{table*}
\begin{figure*}[t]
   \begin{center}
   \includegraphics[width=.95\linewidth]{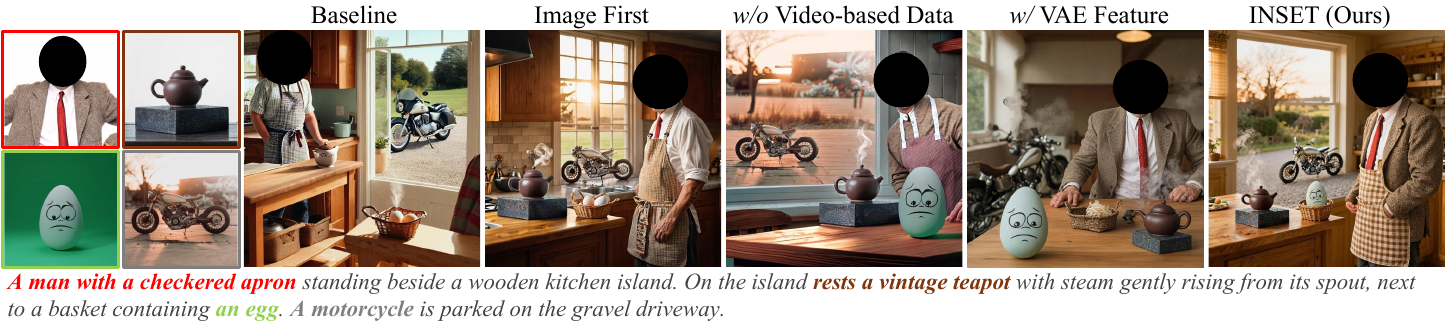}
   \end{center}
   \caption{
   \textbf{Qualitative Ablation Study of Key Model Components.}
   Our interleaved formulation significantly improves attribute binding and context adherence by leveraging semantic locality over the ``Image First'' baseline. Furthermore, incorporating video-based data enhances interactive dynamics, while our ViT-only design avoids the ``image-pasting'' issues and high token overhead typical of VAE-based encoders.
   }
    \label{fig:main_ablation}
\end{figure*}

\subsection{Experimental Setup}
\noindent \textbf{Implementation Details.} 
We initialize $\ours$ from the BAGEL model, fine-tuning all parameters except for the VAE. The model is trained on a composite dataset containing image-based interleaved data, video-based interleaved data, text-guided image editing data, and text-to-image data, with a sampling ratio of $0.2:0.2:0.1:0.5$, respectively. For optimization, we use AdamW with $\beta_1 = 0.9$ and $\beta_2 = 0.95$, setting the learning rate to $2.5 \times 10^{-5}$ for a total of 50k steps. Throughout the training, the maximum image resolution is set to 1024, the sequence length per rank is about 30k, and the diffusion timestep shift is set to $3.0$.

\noindent \textbf{Evaluation Details.}
We utilize Doubao-Seed-1.6 to evaluate both image consistency and text consistency, reporting performance metrics across varying numbers of objects. For the evaluation of baseline methods, we adapt the input format to ensure compatibility. In Specific, while InterleaveBench inherently uses an interleaved format (\textit{e.g.}, ``A [Image 1] dog in [Image 2] park meets [Image 3] cat.''), most baseline methods require reference images to be prepended and indexed. Therefore, we restructure the inputs for these methods by moving all images to the beginning and rewriting the prompts to reference them explicitly via indices.

\subsection{Qualitative Comparisons}
We compare our method with representative open-source models in Fig.~\ref{fig:main_com_open}. 
Experimental results demonstrate that our approach significantly outperforms baselines in both visual consistency and instruction following. 
Specifically, existing methods frequently misalign generated objects or ignore visual inputs entirely, as evidenced by the failure cases of the ``poke ball'' in the third row and the ``anime man'' in the last row. 
Moreover, models such as DreamOmni 2~\cite{xia2025dreamomni2} and Flux-Kontext~\cite{flux2024} exhibit inferior capability in precise attribute binding.
For instance, they fail to render the ``cream-colored sweater'' (second row) or the action to ``relax on a flamingo float'' (last row). 
Finally, comparisons with powerful proprietary models~\cite{gemini2025flashimage,seedream2025seedream,openai2024introducing4o} in Fig.~\ref{fig:main_com_close} further highlight the superiority in maintaining image fidelity in handling complex interleaved instructions, as exemplified by the ``pineapple'' case in the second row.

\subsection{Quantitative Comparisons}
Table~\ref{tab:main_com} presents a comparison of image and text consistency on InterleaveBench against both open-source and proprietary models. 
Experimental results demonstrate that, despite having the fewest parameters, $\ours$ significantly outperforms all open-source methods across all metrics and achieves performance comparable to powerful closed-source models. 
Notably, our advantage over state-of-the-art open-source baselines becomes increasingly pronounced as the number of input images rises. Specifically, in the challenging ``Five Objects'' setting, we lead by substantial margins of $0.29$ in image consistency and $0.24$ in text consistency. 
Furthermore, while our method surpasses closed-source models in image consistency, it slightly lags in text consistency due to the limited capabilities of the underlying text-to-image generation model.

\subsection{Emerging Multimodal Image Editing}
Although trained separately on interleaved instructions and text-guided image editing tasks, $\ours$ successfully integrates these capabilities, leading to the emergence of novel image editing abilities driven by interleaved instructions. Fig.~\ref{fig:main_edit} demonstrates this by showing the model executing edits based on instructions containing both text and specific visual references. A comparison between the "\textit{w/o} Input Images" and "\textit{w/} Input Images" columns reveals that incorporating the input reference images enables highly precise editing. The model faithfully transfers the exact visual characteristics of the specified objects (\textit{e.g.}, specific cap design, branded t-shirt, or robot), rather than relying solely on generic interpretations of the text description.

\subsection{Ablation Studies}
Adopting BAGEL as the baseline, $\ours$ is trained on interleaved data constructed from images and videos, with the exclusion of input image VAE features. This approach significantly enhances the capability to comprehend complex interleaved instructions.
In Fig.~\ref{fig:main_ablation} and Table~\ref{tab:main_ablation}, we investigate the impact of various improvements on the final performance.

\noindent \textbf{Effect of Image Placement.}
To validate our native interleaved formulation, we benchmark it against the standard ``Image First'' approach using identical training data.
As shown in Fig.~\ref{fig:main_ablation} and Table~\ref{tab:main_ablation}, by explicitly positioning visual tokens, we leverage contextual locality to achieve precise object binding.
In contrast, the ``Image First'' paradigm forces the model to resolve difficult long-range dependencies between the text and the prepended images.
Consequently, the baseline often fails to accurately bind attributes to their corresponding targets, or simply neglects specific image inputs entirely.

\noindent \textbf{Effect of Video-based Data.}
We evaluate the impact of integrating video-based interleaved data into our training set. While a model trained solely on image data achieves satisfactory visual consistency and simple spatial alignment, it falters when required to alter object attributes or synthesize interactive dynamics based on text. Adding video data bridges this gap, enabling the model to generate object interactions that are both more natural and semantically accurate.

\noindent \textbf{Effect of VAE Vision Encoder.}
We validate removing the pixel-level VAE encoder by comparing against the standard BAGEL architecture (\textit{i.e.}, \textit{w/} VAE).
Results indicate that VAE latents induce ``image-pasting,'' where the model rigidly copies pixels at the expense of following editing instructions.
Moreover, the excessive token overhead from VAE dilutes the context, causing object omission and inferior consistency compared to our ViT-only design.
\section{Conclusion}
In this paper, we introduced $\ours$, a unified generation model that embeds images as native vocabulary within textual instructions. 
By positioning visual features directly at their corresponding semantic slots, our approach ensures precise object binding and overcomes the limitations of indirect referencing mechanisms. 
To enable this paradigm, we developed a scalable data engine to construct 15M high-quality interleaved samples and validated the model on our proposed InterleaveBench. 
Results show that our method not only excels in complex generation tasks but also naturally generalizes to multimodal image editing. 
We believe that treating visual content as dense and expressive language tokens offers a promising direction for future research to build more intuitive and unified multimodal systems.


\bibliographystyle{plainnat}
\bibliography{main}

\end{document}